# Interpretable Graph Kolmogorov–Arnold Networks for Multi-Cancer Classification and Biomarker Identification using Multi-Omics Data

*Fadi Alharbi[1,2], Nishant Budhiraja[1], Aleksandar Vakanski[1,#], Boyu Zhang[1], Murtada K. Elbashir[2,3], Harshith Guduru[4], and Mohanad Mohammed[5]*

[1] University of Idaho, Department of Computer Science, Moscow, ID 83844, USA

[2] Jouf University, College of Computer and Information Sciences, Department of Computer Science, Sakaka, Aljouf 72441, Saudi Arabia

[3] University of Gezira, Faculty of Mathematical and Computer Sciences, Wad Madani, Sudan

[4] Bentonville High School, Bentonville, AR 72712, USA

[5] University of KwaZulu-Natal, School of Mathematics, Statistics and Computer Science, Pietermaritzburg, Scottsville, 3209, South Africa

[#]Corresponding Authors: Aleksandar Vakanski: vakanski@uidaho.edu

**Abstract**

The integration of heterogeneous multi-omics datasets at a systems level remains a central challenge for developing analytical and computational models in precision cancer diagnostics. This paper introduces Multi-Omics Graph Kolmogorov–Arnold Network (MOGKAN), a deep learning framework that utilizes messenger-RNA, micro-RNA sequences, and DNA methylation samples together with Protein-Protein Interaction (PPI) networks for cancer classification across 31 different cancer types. The proposed approach combines differential gene expression with DESeq2, Linear Models for Microarray (LIMMA), and Least Absolute Shrinkage and Selection Operator (LASSO) regression to reduce multi-omics data dimensionality while preserving relevant biological features. The model architecture is based on the Kolmogorov–Arnold theorem principle and uses trainable univariate functions to enhance interpretability and feature analysis. MOGKAN achieves classification accuracy of 96.28% and exhibits low experimental variability in comparison to related deep learning-based models. The biomarkers identified by MOGKAN were validated as cancer-related markers through Gene Ontology (GO) and Kyoto Encyclopedia of Genes and Genomes (KEGG) enrichment analysis. By integrating multi-omics data with graph-based deep learning, our proposed approach demonstrates robust predictive performance and interpretability with potential to enhance the translation of complex multi-omics data into clinically actionable cancer diagnostics.

## 1. Introduction

Cancer is a highly heterogeneous disease driven by genetic, epigenetic, and transcriptomic alterations. Advances in high-throughput sequencing technologies have enabled the generation of multi-omics datasets, offering deeper insights into the mechanisms underlying cancer development and patient outcomes. Among these, gene expression profiling plays a central role, allowing to monitor gene activity within specific tissues and cell populations and to distinguish cancerous from healthy cells [1]. Messenger RNA (mRNA) levels reflect active gene transcription under specific conditions, providing valuable information on tumor progression and cellular behavior [2], as genes often display altered expression patterns in tumors compared to healthy tissues, revealing key molecular changes associated with cancer [3]. Analyzing such patterns aids in identifying cancer-specific genes and discovering potential biomarkers for early detection. The integration of gene expression data with DNA methylation and microRNA (miRNA) expression profiles has further advanced cancer research [4]. Combining molecular layers uncovers complex regulatory interactions that contribute to tumorigenesis [5]. DNA methylation profiling highlights epigenetic modifications that can silence tumor suppressor genes or activate oncogenes [6], while miRNA expression analysis reveals critical mechanisms of post-transcriptional gene regulation involved in cancer progression [7].

Despite the wealth of information provided by multi-omics data, extracting meaningful insights remains a major challenge due to the high dimensionality, feature heterogeneity, and complexity of genomic structures [8], [9]. Traditional machine learning approaches, such as Support Vector Machines (SVMs) and Random Forests (RF), have shown potential for multi-omics-based cancer classification but often struggle with modeling the complex relationships in high-dimensional datasets and providing interpretable results [10], [11].

Recent advances in deep learning, particularly Graph Neural Networks (GNNs), have demonstrated strong capabilities in capturing complex biological interactions [12], [13]. Unlike conventional models that rely on Euclidean-based representations, GNNs naturally encode the relationships among genes, proteins, and regulatory elements within a graph structure, offering a biologically meaningful approach [14],[15]. The Graph Kolmogorov-Arnold Network (GKAN) represents a significant advancement, whereas by applying Kolmogorov-Arnold representation theory to graph learning, GKAN enhances both model interpretability and flexibility through the use of trainable univariate functions on graph edges [16], [17]. Furthermore, the incorporation of spline-based transformations allows for precise feature extraction and greater transparency, making GKAN particularly well-suited for biomarker discovery in cancer diagnosis.

This article introduces the Multi-Omics Graph Kolmogorov–Arnold Network (MOGKAN), a deep learning framework that integrates graph-based modeling of mRNA, miRNA, and DNA methylation data to classify 31 distinct cancer types. Protein-Protein Interaction (PPI) network information is used for defining the graph structure of MOGKAN. The data preprocessing pipeline combines differential expression analysis, Linear Models for Microarray Analysis (LIMMA) [18], and LASSO regression to extract the most informative multi-omics features. DESeq2 [19] was applied to mRNA data to identify genes exhibiting significant changes in expression levels. For DNA methylation data, LIMMA employs empirical Bayes methods to stabilize variance estimates and improve the detection of differential signals, particularly for genes with low expression levels. This approach enables the identification of differentially methylated regions with high sensitivity and specificity, providing a robust foundation for epigenetic research and the discovery of novel cancer biomarkers.

The primary contributions of this work are as follows:

- Proposed MOGKAN, a novel deep learning framework for cancer classification with inherent feature interpretability through learnable activation functions.
- Constructed a graph-based model integrating a PPI network graph structure with multi-omics data from mRNA, miRNA, and DNA methylation profiles. The combined use of DESeq2, LIMMA, and LASSO enabled the selection of biologically relevant features critical for cancer classification.
- Identified key biomarkers driving cancer progression and validated their functional relevance through Gene Ontology (GO) and Kyoto Encyclopedia of Genes and Genomes (KEGG) pathway analyses.

The rest of this paper is structured as follows. Section 2 discusses related work on graph-based network architectures and Kolmogorov–Arnold networks. Section 3 describes the datasets, preprocessing pipeline, multi-omics data integration, GKAN architecture, and experimental setup. Section 4 presents experimental results and analysis, and biomarker discovery. Section 5 concludes the paper.

## 2. Related Works

Graph Neural Networks (GNNs) have demonstrated significant success in modeling structured data derived from graph-based relationships. While traditional GNNs offer strong predictive performance, they often face two major limitations, related to scalability and interpretability issues. To address these challenges, recent works explored Kolmogorov–Arnold Networks (KAN) as an alternative architecture.

GKAN extends this concept by integrating KAN into graph learning tasks, replacing conventional linear weights with trainable univariate functions, thereby enhancing both model interpretability and flexibility.

Recent studies have increasingly adopted GKAN to improve feature representation and model interpretability in graph-based learning tasks. Zhang et al. [21] introduced GraphKAN, replacing standard activation functions with KAN-based structures to enhance feature extraction, demonstrating superior performance in both node and graph classification tasks compared to traditional GNN architectures. Similarly, Kiamari et al. [22] incorporated spline-based activation functions between graph layers, achieving high performance across a variety of graph network types. Kolmogorov–Arnold Graph Neural Networks (KAGNNs) proposed by Bresson et al. [23] extended message-passing operations using principles from the Kolmogorov–Arnold theorem to improve graph learning. Carlo et al. [24] further refined GKAN by applying spline-based activation functions directly to graph edges, boosting both predictive accuracy and interpretability.

Beyond methodological improvements, several studies have successfully employed GKAN-based architectures in molecular and biomedical domains. Ahmed et al. [25] demonstrated that GKANs can accurately predict small molecule–protein interactions, highlighting their potential in drug discovery. Li et al. [26] developed GNN-SKAN, an architecture combining Swallow-KAN (SKAN) with basic GNNs, achieving state-of-the-art results across multiple molecular datasets. The increasing complexity of biomedical data in multi-omics cancer classification presents a compelling opportunity for GKAN to advance cancer-type prediction as a rapidly evolving research area. GKAN networks are particularly well-suited for applications that demand transparent, interpretable models, as they provide explicit insights into prediction decisions.

Recent approaches such as scRGCL [27] and scMGATGRN [28] combined GNNs with contrastive learning and multiview mechanisms for improved interpretability and learning high-order biological relationships in single-cell transcriptomic data. Furthermore, scAMZI [29] presented attention-based autoencoders as an additional scRNA-seq clustering method, and iCRBP-LKHA [30] used hybrid attention and deep convolutional kernels to predict circRNA-RBP interactions. While these models are employed for cell-type annotation or molecular interaction predictions, MOGKAN is scaled to multi-omics integration and cancer-type classification and enhances both performance and interpretability.

## 3. Materials and Methods

### 3.1. Dataset

This study utilizes data from the Pan-Cancer Atlas database [31], which comprises genomic, transcriptomic, and epigenomic information across a wide range of cancer types. Access to this comprehensive database is facilitated by Genomic Data Commons (GDC), which provides streamlined data retrieval through query tools, such as the TCGAbiolinks package [31]. Developed by the National Cancer Institute, GDC serves as a standardized platform that promotes collaborative cancer research by enabling consistent and cohesive data sharing. For the analysis in this work, the extracted data includes 9,171 DNA methylation samples, 10,668 mRNA expression samples, and 10,465 miRNA expression samples. The number of omics samples for 33 types of cancer and normal tissue is detailed in Table 1 [31]. The compiled multi-omics data resource from the Pan-Cancer Atlas database enables the investigation of complex biological relationships and supports identification of biomarkers for studying tumor biology and clinical outcomes.

**Table 1**. Number of samples per cancer type and normal tissues of TCGA multi-omics data (mRNA, miRNA, and DNA methylation) used in this study.

| **Available Cancer Types** | | **Number of Samples** | | | | | |
|---|---|---|---|---|---|---|---|
| | | **mRNA** | | **miRNA** | | **DNA methylation** | |
| | | **Tumor** | **Normal** | **Tumor** | **Normal** | **Tumor** | **Normal** |
| Uterine corpus endometrial carcinoma | UCEC | 553 | 35 | 545 | 33 | 438 | 46 |
| Adrenocortical carcinoma | ACC | 79 | - | 80 | - | 80 | - |
| Brain lower grade glioma | LGG | 516 | - | 512 | - | 516 | - |
| Sarcoma | SARC | 259 | 2 | 259 | - | 261 | 4 |
| Pancreatic adenocarcinoma | PAAD | 178 | 4 | 178 | 4 | 184 | 10 |
| Esophageal carcinoma | ESCA | 184 | 13 | 186 | 9 | 185 | 16 |
| Prostate adenocarcinoma | PRAD | 501 | 52 | 498 | 52 | 502 | 50 |
| Acute Myeloid Leukemia | LAML | - | | - | | - | |
| Kidney renal clear cell carcinoma | KIRC | 541 | 72 | 544 | 71 | 324 | 160 |
| Pheochromocytoma and Paraganglioma | PCPG | 179 | 3 | 179 | 3 | 179 | 3 |
| Head and Neck squamous cell carcinoma | HNSC | 520 | 44 | 523 | 44 | 528 | 50 |
| Ovarian serous cystadenocarcinoma | OV | 421 | - | 490 | - | 10 | - |
| Glioblastoma multiforme | GBM | 158 | 5 | 5 | - | 140 | 2 |
| Uterine carcinosarcoma | UCS | 57 | - | 57 | - | 57 | - |
| Mesothelioma | MESO | 87 | - | 87 | - | 87 | - |
| Testicular germ cell tumors | TGCT | 150 | - | 150 | - | 150 | - |
| Kidney chromophobe | KICH | 66 | 25 | 66 | 25 | 66 | - |
| Rectum adenocarcinoma | READ | 166 | 10 | 161 | 3 | 98 | 7 |
| Uveal melanoma | UVM | 80 | - | 80 | - | 80 | - |
| Thyroid carcinoma | THCA | 505 | 59 | 506 | 59 | 507 | 56 |
| Liver hepatocellular carcinoma | LIHC | 371 | 50 | 372 | 50 | 377 | 50 |
| Thymoma | THYM | 120 | 2 | 124 | 2 | 124 | 2 |
| Cholangiocarcinoma | CHOL | 35 | 9 | 36 | 9 | 36 | 9 |
| Lymphoid neoplasm diffuses large B-cell lymphoma | DLBC | 48 | - | 47 | - | 48 | - |
| Kidney renal papillary cell carcinoma | KIRP | 290 | 32 | 291 | 34 | 275 | 45 |
| Bladder urothelial carcinoma | BLCA | 412 | 19 | 417 | 19 | 418 | 21 |
| Skin cutaneous melanoma | SKCM | 103 | 1 | 97 | 2 | 104 | 2 |
| Lung squamous cell carcinoma | LUSC | 502 | 51 | 478 | 45 | 370 | 42 |
| Stomach adenocarcinoma | STAD | 412 | 36 | 446 | 45 | 395 | 2 |
| Lung adenocarcinoma | LUAD | 539 | 59 | 519 | 46 | 473 | 32 |
| Colon adenocarcinoma | COAD | 481 | 41 | 455 | 8 | 312 | 38 |
| Cervical squamous cell carcinoma and endocervical adenocarcinoma | CESC | 304 | 3 | 307 | 3 | 307 | 3 |
| Breast invasive carcinoma | BRCA | 1,111 | 113 | 1,096 | 104 | 793 | 97 |
| **Tumor and Normal** | | **9,928** | **740** | **9,791** | **670** | **8,424** | **747** |
| **Total** | | **10,668** | | **10,461** | | **9,171** | |

### 3.2. Data Preprocessing

The data preprocessing pipeline in our work integrates dimensionality reduction techniques and feature selection methods to handle high-dimensional data and identify biologically relevant features in omics data. Specifically, we employed LIMMA and differential gene expression (DGE) analysis for feature selection. DGE analysis based on DESeq2 was applied to mRNA expression data, employing a negative binomial model to detect genes with significant expression changes [32], [33]. LIMMA was used to analyze DNA methylation data and identify differentially methylated CpG sites [33]. To further reduce data dimensionality, we applied LASSO regression to mRNA and DNA methylation data [34]. The flow chart for data preprocessing is depicted in Figure 1, with the following sections outlining the phases in the data processing workflow.

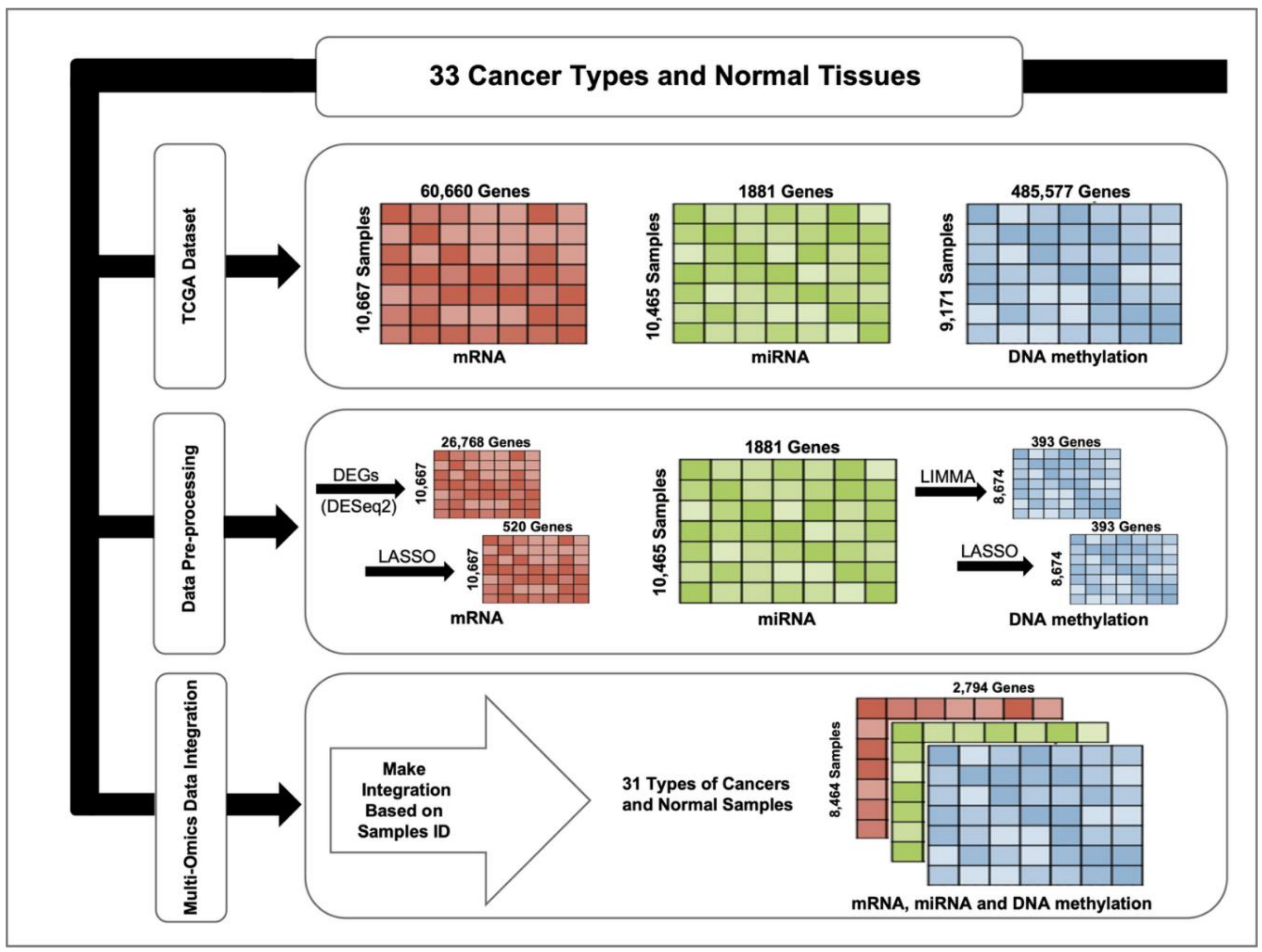

**Figure 1.** Flow chart for data preprocessing.

### *3.2.1. Differential Gene Expression (DGE) Analysis*

In genomics, DGE profiling is frequently used to compare the expression levels of genes in a particular organism under various settings or conditions (e.g., normal versus cancer, treatment against control, etc.) [35]. The analysis helps elucidate gene regulation mechanisms, environmental influences on gene activity, and a variety of other underlying biological processes. In our investigation, we employed DESeq2 to perform differential gene expression analysis on the mRNA data. DESeq2 models gene-level count data using a negative binomial distribution, which effectively accounts for both biological variability and overdispersion. We assessed the statistical significance of gene expression changes using the Wald test, based on *p*-values derived from the Wald statistic to evaluate whether the estimated log fold changes are significant. To identify genes potentially relevant to the biological processes under study, we applied a *p*-value threshold of 0.001.

### *3.2.2. LIMMA*

For differential methylation analysis, we applied the LIMMA technique by fitting a linear model to the methylation levels of CpG sites as a function of experimental sample groups [36]. The initial dataset derived from the Human Methylation 450K (HM450) array included 485,577 features across 9,171 samples [37] (Table 2). Using LIMMA, we identified CpG sites that are significantly differentially methylated in tumor samples compared to normal controls. For each CpG site, LIMMA computes a moderated $t$-statistic and an effect size that captures the relative methylation differences between groups. The corresponding $p$-value indicates the statistical significance of each comparison. After applying a $p$-value cutoff of 0.05, the number of CpG features was reduced to 139,321, representing the most notable methylation alterations associated with the disease state.

*3.2.3. LASSO Regression*

Lasso Regression is a linear regression technique that incorporates $L_1$ norm regularization to enhance model performance and reduce overfitting. The algorithm minimizes the sum of squared residuals while imposing penalties proportional to the absolute values of model coefficients. The enforcement of such penalty encourages sparsity by shrinking some coefficients to zero, effectively performing feature selection by eliminating less important variables. The Lasso objective function is given by the following equation:

$$\min_{\beta} \sum_{i=1}^{n} \left( \gamma_i - \sum_{j=1}^{p} \chi_{ij} \beta_j \right)^2 + \lambda \sum_{j=1}^{p} |\beta_j| \quad (1)$$

where $\gamma_i$ is the observed response variable for the $i$th sample, $\chi_{ij}$ denotes the feature values, $\beta_j$ are the regression coefficients, $\lambda$ is the regularization parameter that controls the degree of sparsity, and $n$ is the number of samples.

### 3.3. Multi-Omics Data Integration

To integrate mRNA (RNA-Seq), miRNA, and DNA methylation data into unified records, we used sample IDs as the linkage element. An inner join operation was performed on the common sample IDs across the three omics datasets, retaining only those samples that have complete data for all modalities. Cancer types lacking any of the omics layers were excluded from further analysis. Notably, two cancer types LAML and GCT (Table 1) were excluded from further analysis due to missing RNA-Seq and miRNA data, respectively. The final integrated dataset contains 8,464 samples spanning 31 cancer types and corresponding normal tissues, encompassing a total of 2,794 omics features (as summarized in Table 2).

**Table 2**. Pipeline for data processing.

| Datatype | mRNA | miRNA | DNA Methylation |
|---|---|---|---|
| **Original Features** | 60,660 | 1881 | 485,577 |
| **Differentially Expressed Analysis** | 26,768 | - | - |
| **LIMMA Model (Selected Features)** | - | - | 139,321 |
| **LASSO Regression Model (Selected Features)** | 520 | - | 393 |
| **All Tumor Samples and Normal** | 10,668 | 10,465 | 9,171 |
| **Unique Tumor Samples and Normal** | 10,667 | 10,465 | 8,674 |
| **Common Samples and Features** | 8,464 Samples and 2,794 Features | | |

Beside the used early integration strategy where multi-omics modalities are concatenated before passing them to a graph-based model, other integration strategies have been applied in prior works. Picard et al. [38] described multi-omics integration under five integration types: early, mixed, intermediate, late, and hierarchical integration. These strategies have trade-offs in relation to the complexity of the model and pertaining to the interpretability and capability of preserving modality-specific signals. Although simple and popular, early integration methods might not be ideal in dealing with heterogeneity and differences in feature dimensionalities between layers of omics data. Conversely, mixed or intermediate integration have the ability to maintain modality-specific structures, as well as to enable more flexible modeling pipelines. Alternative strategies, like late or hierarchical integration, where each type of omics has its own encoder and they are subsequently fused, may also enhance interpretability and predictive power. Different integration strategies will be investigated in future versions of our framework.

### 3.4. Graph Kolmogorov–Arnold Networks (GKAN)

GKAN represents a neural architecture that extends the Kolmogorov–Arnold Representation Theorem to graph-structured data, offering an alternative to traditional GNNs. Namely, unlike traditional GNNs which rely on message passing, GKAN utilizes functional decomposition to model interactions within graphs. By decomposing node and edge relationships into hierarchical, learnable transformations, GKAN can effectively capture long-range dependencies supported by Kolmogorov-Arnold representation while addressing the over-smoothing problem by using learned edge activation that often affects GNNs [39]. GKAN expresses multi-dimensional functions as summations of nonlinear one-dimensional functions, enabling the adaptive transformation of node embeddings based on information from neighboring nodes. This is grounded in the Kolmogorov-Arnold theorem, which states that any continuous multivariate function $f\colon \mathbb{R}^d \rightarrow \mathbb{R}$ can be decomposed as:

$$f(x_1, x_2, \dots, x_d) = \sum_{q=1}^{2d+1} g_q\left(\sum_{p=1}^{d} h_{q,p}\left(x_p\right)\right) \tag{2}$$

where $g_q$ and $h_{q,p}$ are learnable nonlinear functions, $x_p$ denotes the input features, and $d$ is the input feature dimension.

For a given graph $G = (V, E)$, where $V$ is the set of nodes and $E$ is the set of edges, the node features $h_v^{(l)}$ at a layer $l$ are updated using:

$$h_v^{(l+1)} = \sum_{q=1}^{2d+1} g_q\left(\sum_{u \in N(v)} h_{q,p}\left(h_u^{(l)}\right)\right) \tag{3}$$

In (3), $h_v^{(l)}$ denotes the feature representation of node $v$ at layer $l$, $\mathcal{N}(v)$ represents the set of neighboring nodes of $v$, and $g_q$ and $h_{q,p}$ are trainable transformation functions applied to graph features. After several layers of hierarchical transformations, the final node representation is obtained as:

$$\hat{y}_v = \sigma\left(\sum_{q=1}^{2d+1} g_q\left(\sum_{p=1}^{d} h_{q,p}\left(h_v^{(L)}\right)\right)\right) \tag{4}$$

where $\hat{y}_v$ is the predicted class or regression output for node $v$, $\sigma$ is an activation function such as softmax (for classification) or sigmoid (for binary prediction), and $L$ is the total number of layers in the network.

### 3.5. Graph Structure

Protein-protein interactions are fundamental to biological systems, as they represent physical contacts or functional relationships between two or more protein molecules. The interactions play a central role in regulating cellular processes. In this study, we constructed a PPI network using the STRING database, which integrates both experimentally validated and computationally predicted protein interaction data [40]. STRING aggregates data from diverse biological sources, including high-throughput experimental assays, curated pathway databases, co-expression analyses, and text-mined associations extracted from scientific literature [41]. To enhance the reliability of interaction data, STRING assigns confidence scores to each interaction based on the strength and consistency of supporting evidence, thereby improving the robustness of biological network analyses [39].

For our analysis, we focused on constructing a disease-specific protein network that connects proteins associated with multi-omics-derived genes, including mRNA, miRNA, and DNA methylation profiles. We utilized the STRING API (version 11.5) to automatically retrieve Homo sapiens (NCBI Taxonomy ID: 9606) protein interaction data. The results in TSV format were accompanied by confidence scores derived from co-expression data, experimental findings, curated databases, and literature mining. The resulting graph based on PPI networks was afterward used as input to the MOGKAN model, enabling biological graph representations that support cancer classification and biomarker discovery.

### 3.6. Experimental Setting

The pipeline of the proposed framework is illustrated in Figure 2. To construct the graph structure, we utilized a PPI-based edge index, derived by identifying highly interactive proteins within the PPI network. The selection is based on protein frequency counts, where only proteins appearing at least 200 times in the dataset are retained, ensuring the inclusion of biologically significant hub proteins with prominent roles in cellular function and interaction networks. This filtering step also aids in adjusting the graph's layout and the amount of data it contains. Excluding weakly connected hubs removes isolated nodes from the graph, resulting in a more cohesive and interpretable graph structure.

The layers in the MOGKAN architecture are depicted in Figure 2. Multi-omics data for each sample, including mRNA, miRNA, and DNA methylation, are integrated into a unified feature vector that is assigned to the corresponding nodes in the graph. Information is propagated across the network using GATConv layers [42], which dynamically assign weights to neighboring nodes based on an attention mechanism learned during training. This allows the model to prioritize the most informative interactions by computing relevance scores for each neighbor. Through multiple GAT layers, the network iteratively refines node representations by aggregating attention-weighted messages, effectively capturing both biological signals and meaningful connectivity patterns. The resulting node embeddings encode local gene-specific characteristics while also reflecting the broader structure of PPI networks.

To fine-tune hyperparameters we employed a grid search strategy for learning rate, weight decay, dropout rate, and the number of attention heads. Model training was performed using the Adam optimizer over 100 epochs. The results of a grid search for the MOGKAN model are presented in Table 3, which shows the top 10 hyperparameter combinations that resulted in the highest mean accuracy and F1-score during grid search optimization of the model on multi-omics data. Several different arrangements were tried by altering the main architecture and training details like the hidden dimension size, number of attention heads, the hidden layers, dropout rate, learning rate, and L2 regularization strength. The model achieved consistently high performance across multiple settings, obtaining mean accuracies of 96.1% and F1-scores exceeding 95%. The best achieved results were in the configuration using a hidden dimension of 2048, four attention heads, two hidden layers, a dropout rate of 0.2, a learning rate of 0.0001, an L2 penalty of 0.0001, and yielded an accuracy of 96.17% and an F1-score of 95.12%.

As depicted in Figure 2, the framework employs 5-fold cross-validation ensuring robust evaluation across different data splits. Data from multi-omics is processed by two consecutive GAT layers. The first GAT multiplies feature vectors by four attention heads, each with 2048-dimensional information, followed by a second GAT that sums the outputs and a LeakyReLU is used as its activation. The output is then passed through three Kolmogorov–Arnold Network (KAN) layers that apply nested nonlinear transformations ($\psi \rightarrow \tanh \rightarrow \varphi \rightarrow \tanh$), each followed by batch normalization, LeakyReLU activation, and dropout for regularization. Feature dimensions are progressively reduced from the initial hidden size to 1024 and then to 512. Lastly, a linear classifier takes each processed set of attributes and converts them to 32 categories for the different cancer types. This hybrid design enables MOGKAN to capture both topological dependencies and complex nonlinear patterns in multi-omics cancer data.

**Table 3.** Grid search results for hyperparameter optimization of the MOGKAN model on multi-omics data.

| Hidden_Dim | Gat_Heads | Hidden_Layers | Dropout_Rate | Learning_rate | L2_penalty | Mean_Accuracy | Mean_F1 |
|---|---|---|---|---|---|---|---|
| 2048 | 4 | 2 | 0.2 | 0.0010 | 0.0001 | 0.9618 | 0.9530 |
| 1024 | 8 | 2 | 0.2 | 0.0010 | 0.0001 | 0.9610 | 0.9514 |
| 2048 | 8 | 2 | 0.3 | 0.0001 | 0.0001 | 0.9616 | 0.9513 |
| 1024 | 4 | 2 | 0.3 | 0.0010 | 0.0001 | 0.9604 | 0.9512 |
| 2048 | 4 | 2 | 0.2 | 0.0001 | 0.0001 | 0.9617 | 0.9512 |
| 2048 | 8 | 2 | 0.2 | 0.0001 | 0.0001 | 0.9612 | 0.9508 |
| 1024 | 8 | 3 | 0.2 | 0.0010 | 0.0000 | 0.9608 | 0.9508 |
| 1024 | 8 | 2 | 0.3 | 0.0010 | 0.0001 | 0.9610 | 0.9507 |
| 2048 | 8 | 2 | 0.3 | 0.0001 | 0.0000 | 0.9616 | 0.9507 |
| 2048 | 8 | 3 | 0.2 | 0.0001 | 0.0001 | 0.9608 | 0.9505 |

Performance evaluation was conducted using standard classification metrics, including accuracy, precision, recall, and F1-score, averaged across multiple folds of cross-validation. To interpret the model's predictions, feature importance was assessed based on activations in the model's GAT first layer. For every attention head, the model learns a set of coefficients to determine the importance of nearby genes (or nodes) when aggregating data. To find out the importance of each gene, we averaged the attention scores given to each gene across the first layer of the GAT. Instead of using just one "head," this approach captures a comprehensive view of the model's attention mechanism. After the training process, we obtained the average attention weights assigned to each node across all samples and heads. Genes that consistently received high attention scores across different samples were considered more influential, as they contributed more significantly to feature propagation and decision-making within the graph. The most influential features were mapped to their corresponding genes using a BioMart query, providing enhanced biological context and insight into their relevance in cancer classification.

### 3.7. Performance Metrics

For model evaluation, we applied standard performance metrics for multi-class classification tasks including accuracy, precision, recall, and F1-score. The model accuracy serves as a measure of the overall correctness defined through the following equitation:

$$Accuracy = \frac{\sum_{i=1}^{N} TP_i}{\sum_{i=1}^{N} (TP_i + TN_i + FP_i + FN_i)} \tag{5}$$

Macro-averaging enables calculation of precision, recall, and F1-score per class before computing their collective average without preference to any class, as follows.

$$Macro\ Precision = \frac{1}{N} \sum_{i=1}^{N} \frac{TP_i}{TP_i + FP_i} \tag{6}$$

$$Macro\ Recall = \frac{1}{N} \sum_{i=1}^{N} \frac{TP_i}{TP_i + FN_i} \tag{7}$$

$$Macro\ F1-score = \frac{1}{N}\sum_{i=1}^{N}\frac{2 \times Precision_i \times Recall_i}{Precision_i + Recall_i} \tag{8}$$

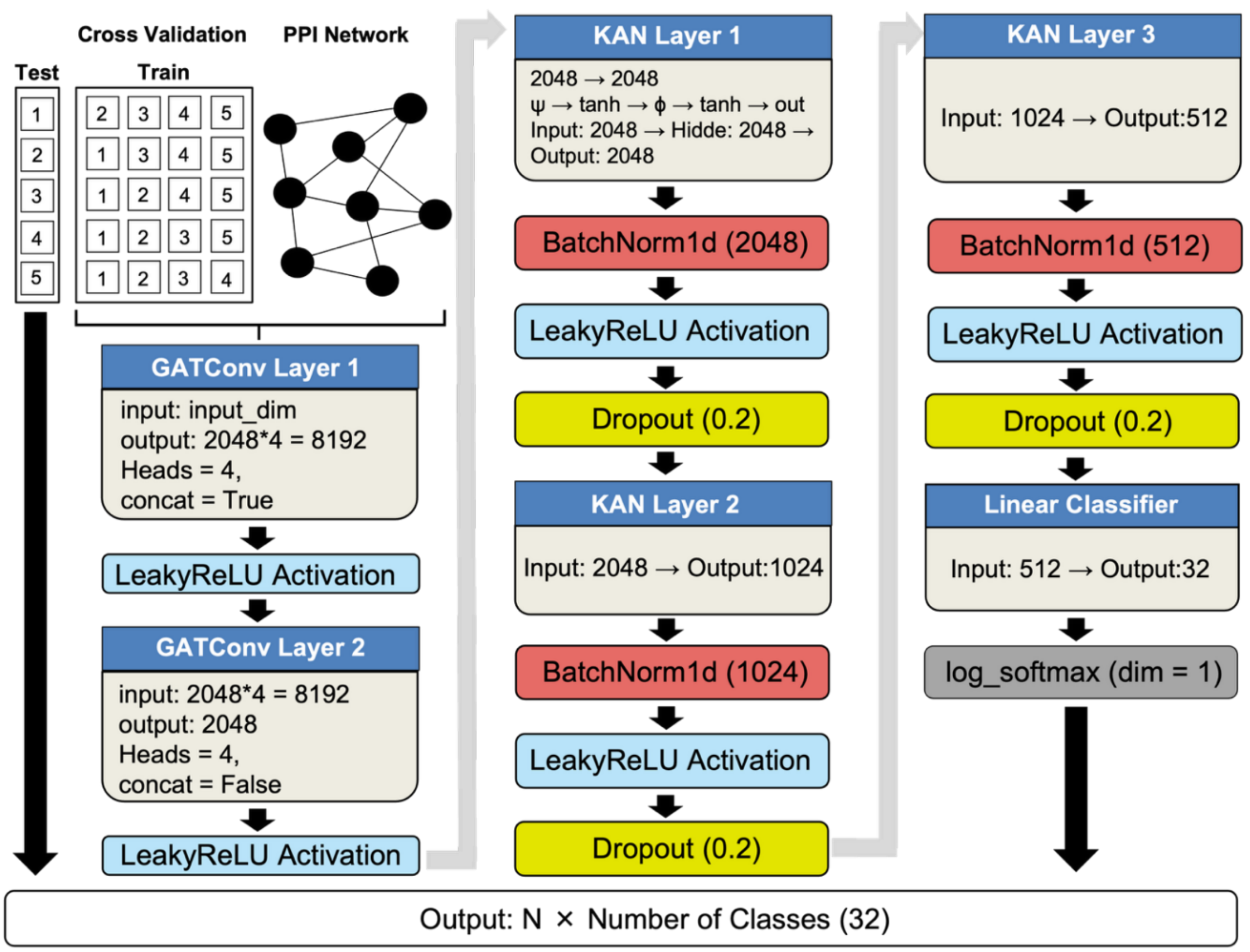

**Figure 2**. The architecture of the Multi-Omics Graph Kolmogorov-Arnold Network (MOGKAN) for multiclass cancer classification.

## 4. Results and Discussion

The performance of the proposed MOGKAN framework evaluated using 5-fold cross-validation is summarized in Table 4. MOGKAN achieved a classification accuracy of 96.28% across 32 cancer types by integrating mRNA, miRNA, and DNA methylation data. The results demonstrate performance improvement ranging from 1.58% to 7.30% compared to related works employing deep learning architectures based on Convolutional Neural Network (CNN), Graph Convolutional Neural Network (GCNN), and Graph Transformer Network (GTN). In particular, Mostavi et al. [43] achieved 95.70% accuracy using a CNN-based model, Ramirez et al. [44] reported 94.61% accuracy with a GCNN-PPI approach, whereas Kaczmarek et al. [45] implemented a GTN model that achieved 93.56% accuracy. Moreover, MOGKAN exhibits improved reliability as evidenced by its low standard deviation across the multiple folds of ±0.0035, which stands out against the variability in the results of related works.

**Table 4.** Experimental results for related deep learning methods for cancer classification with omics data.

| Authors | Models | Pan-Cancer | Multi-Omics Data type | | | Accuracy |
|---|---|---|---|---|---|---|
| | | | mRNA | miRNA | DNA methylation | |
| Mostavi et al., 2020 [43] | 1D-CNN<br>2D-Vanilla-CNN<br>2D-Hybrid-CNN | 34 Classes | √ | - | - | 95.50 ± 0.1<br>94.87 ± 0.4<br>95.70 ± **1.0** |
| Ramirez | GCNN-PPI graph<br>GCNN-PPI + singleton graph | 34 Classes | √ | - | - | 88.98 ± **0.9**<br>94.61 ± **1.0** |
| et al., 2020 [44] | | | | | | |
| Kaczmarek et al., 2022 [45] | GTN | 12 Classes | √ | √ | - | 93.56 **± 0.9** |
| **Proposed MOGKAN-PPI graph** | | 32 Classes | √ | √ | √ | **96.28 ± 0.0035** |

Table 5 presents the experimental evaluation of MOGKAN with single-omics and multi-omics data for classification of 31 cancer types and normal tissues. Among the single-omics inputs, the model trained with mRNA data achieved the highest accuracy of 0.9562 along with 0.9524 precision, 0.9357 recall, and 0.9414 F1-score. The multi-omics MOGKAN model trained with combined DNA methylation and miRNA data performed similarly to the single-omics models, although the results remain slightly below top performance levels. The combined use of mRNA, DNA methylation, and miRNA data resulted in the most effective performance including 0.9628 accuracy, 0.9582 precision, 0.9445 recall, and 0.9489 F1-score. These results confirm that the integration of multiple omics modalities enhances the performance across all evaluation metrics.

**Table 5.** Experimental results for the proposed MOGKAN approach with single-omics and multi-omics data.

| Data Types | Multi-Omics Data Type | Accuracy | Precision | Recall | F1 Score |
|---|---|---|---|---|---|
| Single Omics Data | mRNA or RNA-Seq | **0.9562 ± 0.0029** | 0.9524± 0.0074 | 0.9357± 0.0094 | 0.9414 ± 0.0072 |
| Single Omics Data | miRNA | 0.9545 ± 0.0029 | 0.9497 ± 0.0062 | 0.9307 ± 0.0091 | 0.9373 ± 0.0069 |
| Single Omics Data | DNA methylation | 0.9551 ± 0.0024 | 0.9437 ± 0.0121 | 0.9263 ± 0.0151 | 0.9324 ± 0.0131 |
| Multi-Omics Data | mRNA or RNA-Seq and miRNA | 0.9553 ± 0.0037 | 0.9515 ± 0.0087 | 0.9337 ± 0.0101 | 0.9396 ± 0.0069 |
| Multi-Omics Data | mRNA or RNA-Seq and DNA methylation | 0.9546 ± 0.0036 | 0.9519 ± 0.0065 | 0.9312 ± 0.0091 | 0.9387 ± 0.0073 |
| Multi-Omics Data | miRNA and DNA methylation | 0.9548 ± 0.0047 | 0.9505 ± 0.0082 | 0.9319 ± 0.0099 | 0.9383 ± 0.0084 |
| **Multi-Omics Data** | mRNA or RNA-Seq, miRNA and DNA methylation | **0.9628 ± 0.0035** | **0.9582 ± 0.0082** | **0.9445 ± 0.0124** | **0.9489 ± 0.0087** |

Notably, the removal of miRNA data resulted in the most significant drop in model performance, compared to excluding either mRNA or DNA methylation data. While mRNA is typically considered central to assessing gene expression and tumor signals, several biological and technical factors explain the impact of miRNA. Biologically, miRNAs play a critical role in post-transcriptional gene regulation, often acting as oncogenes or tumor suppressors. Dysregulation of miRNAs can influence the expression of numerous genes simultaneously, making them key indicators of cancer progression and subtype differentiation. Additionally, miRNAs are fewer in number and exhibit more stable expression patterns than mRNAs, which may help the model identify more robust and generalizable patterns. From a technical standpoint, the GraphKAN model's attention mechanism assigns dynamic weights to input features. The stronger influence of miRNA features suggests that they contributed more prominently during training, allowing the model to focus on their informative value more effectively.

To further evaluate the generalization capability of MOGKAN, we performed a type-blind evaluation that withheld TCGA-LUAD, TCGA-LUSC, and TCGA-PRAD during training and used them only for testing. Table 6 demonstrates that the model maintains good predictive performance even in this rigorous evaluation. With the TCGA-PRAD cancer type the model achieved an accuracy of 0.9847, precision of 0.9973, and recall of 0.9620, indicating excellent generalization. With the TCGA-LUAD cancer type the model also showed robust performance with an accuracy of 0.9590, and an F1 score of 0.9393. Although the model performance on TCGA-LUSC was comparatively lower (accuracy = 0.9237, recall = 0.7945), it still reflects effective model generalization under type-blind conditions.

Table 7 lists the top 10 biomarkers identified by the MOGKAN framework based on feature importance. The importance of each feature was quantified using the absolute sum of weights from the model’s linear transformation layer, enabling a discriminative selection of key biomarkers. BioMart was employed to map the features to their corresponding gene identifiers, enhancing biological interpretability. The ten identified biomarkers MCL1, LINC01410, GALNT6, MAML3, ITGB3, LINC01090, PKDCC, PCAT14, KIF16B, and PITPNM3 showed cancer-specific functional patterns that align with known mechanisms of carcinogenesis. For instance, MCL1 was reported to contribute to therapy resistance in breast cancer by regulating mitochondrial oxidative phosphorylation activity (PMID: 28978427) [46]. GALNT6 and PITPNM3 have been identified as dual-function proteins promoting both epithelial-mesenchymal transition and immune evasion (PMIDs: 39245709, 21481794) [47], [48]. Several long non-coding RNAs, including LINC01410 (PMID: 32104067), LINC01090 (PMID: 34550610), and PCAT14 (PMID: 35003397), were found to participate in ceRNA regulatory networks. Notably, PCAT14 exhibited the highest diagnostic precision for prostate cancer [49], [50], [51]. Furthermore, ITGB3 and KIF16B were implicated in extracellular vesicle-mediated communication, with evidence supporting their role as potential biomarkers for metastatic colorectal cancer (PMIDs: 37040507, 35487942) [52], [53]. The activity of MAML3 is regulated by hypoxia-inducible factors, activating Hedgehog (HH) and NOTCH pathways in gallbladder cancer (GBC), thereby promoting tumor growth, migration, and invasion, while also enhancing sensitivity to gemcitabine (PMID: 37351966) [54]. Lastly, PKDCC has been linked to non–small cell lung cancer progression (PMID: 35847849) [55].

**Table 6.** Per-class performance metrics under type-blind evaluation settings.

| Cancer Types | Accuracy | Precision | Recall | F1 Score |
|---|---|---|---|---|
| **TCGA-LUAD** | 0.9590 | 0.9596 | 0.9198 | 0.9393 |
| **TCGA-LUSC** | 0.9237 | 0.9206 | 0.7945 | 0.8529 |
| **TCGA-PRAD** | 0.9847 | 0.9973 | 0.9620 | 0.9793 |

**Table 7.** Top ten identified pan-cancer biomarkers and supporting evidence.

| Gene Stable ID | Gene Name | Evidence |
|---|---|---|
| **ENSG00000143384** | MCL1 | PMID: 28978427 |
| **ENSG00000238113** | LINC01410 | PMID: 32104067 |
| **ENSG00000139629** | GALNT6 | PMID: 39245709 |
| **ENSG00000196782** | MAML3 | PMID: 37351966 |
| **ENSG00000259207** | ITGB3 | PMID: 37040507 |
| **ENSG00000231689** | LINC01090 | PMID: 34550610 |
| **ENSG00000162878** | PKDCC | PMID: 35847849 |
| **ENSG00000280623** | PCAT14 | PMID: 35003397 |
| **ENSG00000089177** | KIF16B | PMID: 35487942 |
| **ENSG00000091622** | PITPNM3 | PMID: 21481794 |

### 4.1. GO and KEGG Analysis

Figure 3 and Figure 4 depict the top 10 enriched Gene Ontology (GO) terms identified through MOGKAN analysis, highlighting molecular systems that contribute to multi-cancer classification. Figure 3 represents the top 10 enriched Gene Ontology (GO) terms for biological processes. Each horizontal bar corresponds to a GO term, with the length of the bar indicating the degree of enrichment, where the taller the bar the more enriched GO terms are. Our GO analysis identified the top 10 terms that our genes are associated with them, including positive regulation of respiratory burst, regulation of respiratory burst and apoptotic cell clearance, which highlight biological processes that are significantly overrepresented in the analyzed top 50 gene set. These GO terms reflect the role of reactive oxygen species (ROS) in modulating tumor microenvironment dynamics and immunotherapy outcomes [56]. Overall, the enriched GO terms validate the biological relevance of MOGKAN's graph-based integration of multi-omics data, reinforcing its ability to uncover functionally significant pathways involved in cancer development and classification.

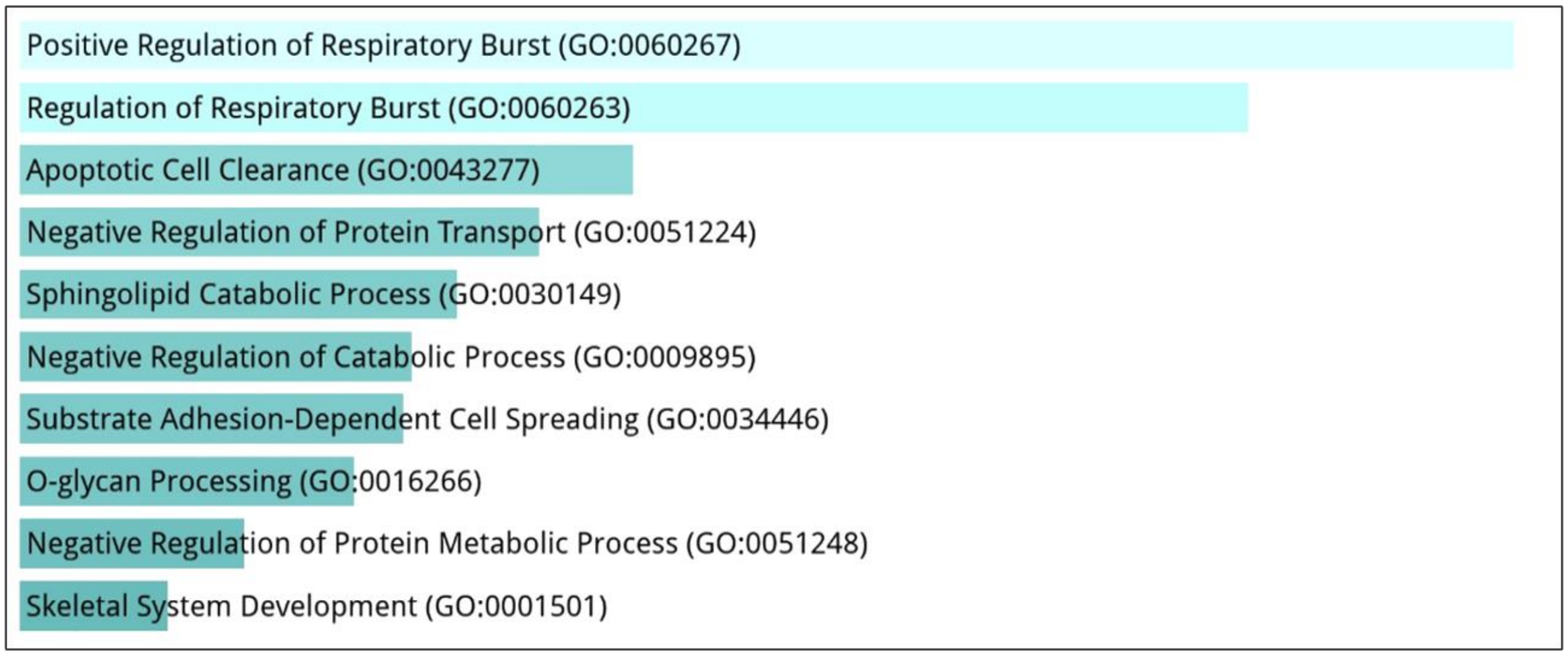

**Figure 3**. Gene ontology enrichment (biological process) of top 50 multi-cancer biomarkers identified by Graph Kolmogorov-Arnold Networks.

Figure 4 depicts the top 10 significantly enriched Gene Ontology (GO) molecular functions, primarily focused on lipid binding and ion channel regulation, suggesting roles in cellular signaling and membrane dynamics. The prominence of phosphatidylinositol (PI) binding terms—such as Phosphatidylinositol-3,5-Bisphosphate Binding (GO:0080025) and Phosphatidylinositol-3-Phosphate Binding (GO:0032266)—indicates involvement in phosphoinositide signaling, a pathway critical for membrane trafficking, autophagy, and cell survival. These findings align with established research indicating that dysregulation of phosphatidylinositol metabolism is prevalent across various cancers, as it activates the PI3K–AKT–mTOR signaling pathway [57] and contributes to treatment resistance [55].

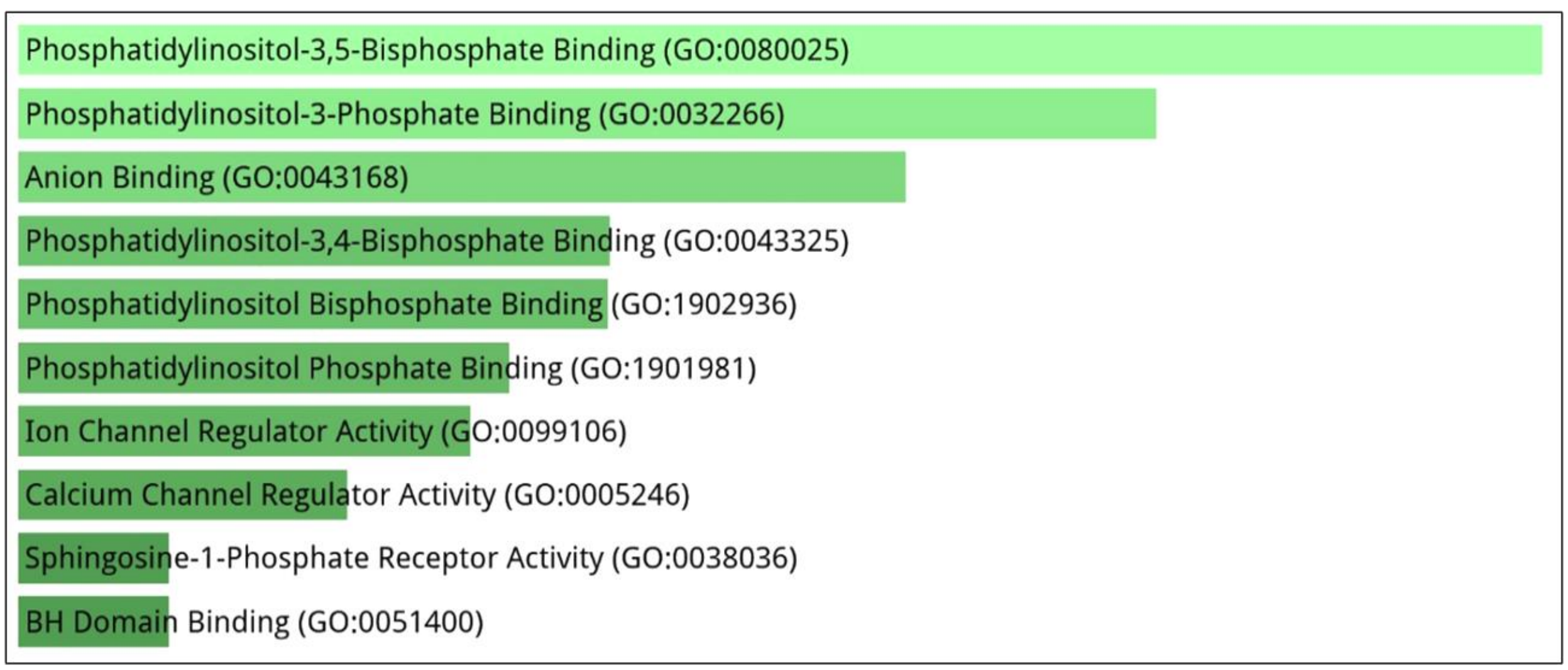

**Figure 4.** Gene ontology enrichment (molecular function) of top 50 multi-cancer biomarkers identified by Graph Kolmogorov-Arnold Networks.

Figure 5 presents the cancer-related gene set enriched KEGG pathways [58–60], ranked by statistical significance using $-\log_{10}$(p-value) scores. The "Mucin type O-glycan biosynthesis" pathway emerged as the most significantly enriched, highlighting its role in altering glycosylation patterns on tumor cells, a known contributor to cancer progression [61], [62]. Closely following is the "Sphingolipid metabolism" pathway, which supports tumor cell survival and resistance to therapeutic agents [63]. The "Prolactin signaling pathway" ranks next in significance, particularly relevant for its involvement in breast cancer regulation [64]. Additionally, approximately 15% of all cancer-associated genes in the dataset are mapped to the "PI3K-Akt signaling pathway", which serves as a central regulator of cellular proliferation and apoptosis yet remains frequently abnormal in cancer development [65]. Further down the ranking, the "Rap1 signaling pathway" was identified, underscoring its role in cell adhesion and metastasis, which aligns with the invasive phenotypes observed in several cancers [66]. The pathways "Aldosterone-regulated sodium reabsorption", "Type I diabetes mellitus", and "Maturity onset diabetes of the young" were also enriched, suggesting shared metabolic disruptions between cancer and diabetes [67], [68]. While several pathways show moderate enrichment with $-\log_{10}$(p-values) around 0.6, "Mucin type O-glycan biosynthesis" exhibits the strongest enrichment signal. Collectively, these results reveal biological pathways that describe mechanisms of cancer development, particularly those related to glycosylation events, lipid metabolism, and growth factor signaling.

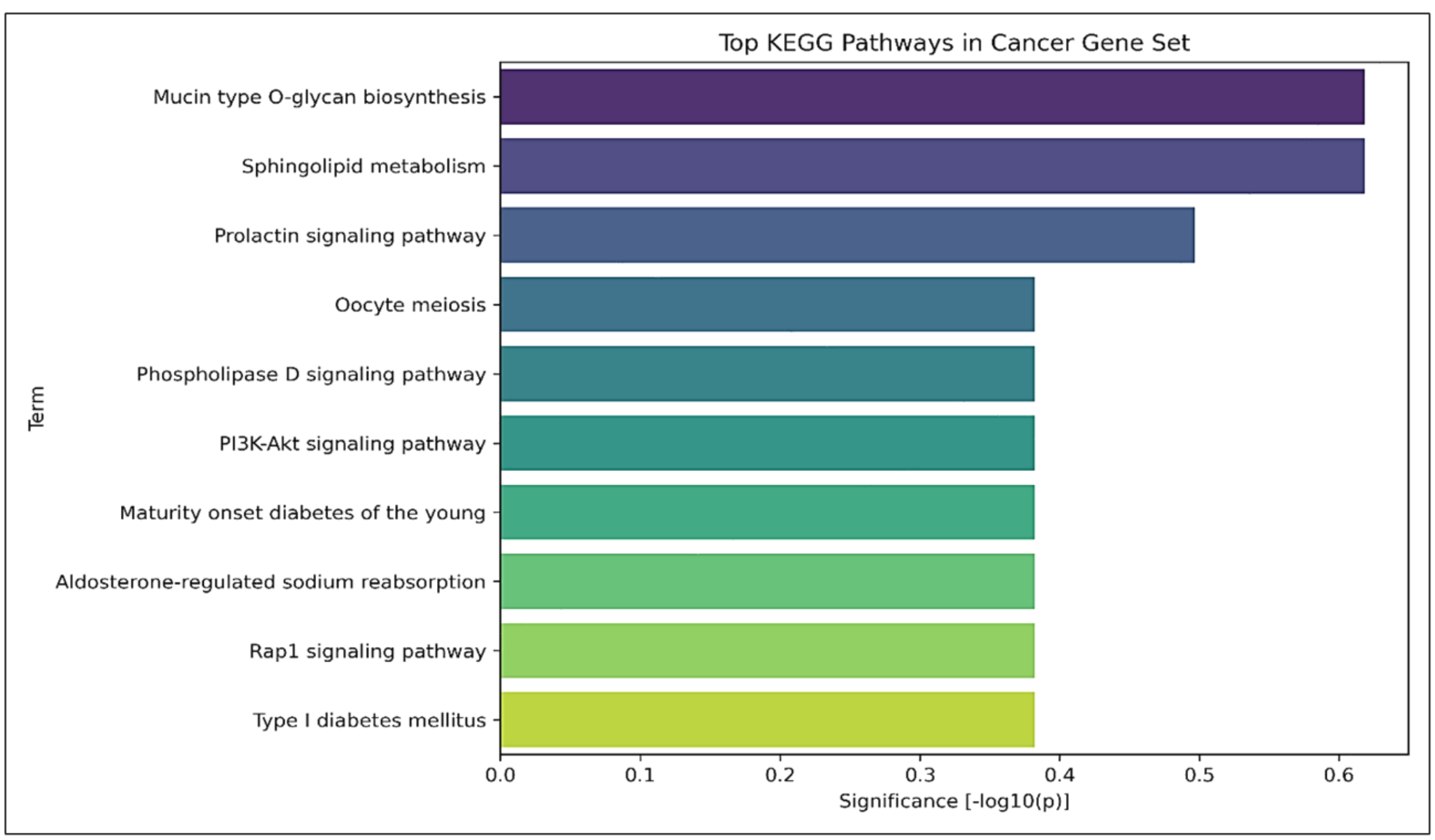

**Figure 5**. Significantly enriched KEGG pathways [58–60] associated with cancer.

The presence of domain-specific deep learning applications to a variety of biological questions underpins the importance of focused, explainable models. As an illustration, in NSCLC ferroptosis-associated lncRNAs were reported to be strong prognosticators and predictors of immunotherapy efficacy [69]. On the same note, deep learning has been used to rebuild the features of protein transport [70] and diagnose cleft lip and palate through imaging-based ML models [71]. All of these studies demonstrate the increasing range of applications of interpretable models in clinical diagnostic, which is consistent with the goal of MOGKAN to discover biologically relevant cancer biomarkers using graph-based learning.

### 4.2. Limitations and Future Work

The proposed MOGKAN framework has several limitations. First, it relies on static PPI network information data from the STRING database, which may lack dynamic or condition-specific protein interactions. Incorporating tissue-specific or context-aware interaction networks could strengthen the biological relevance of the constructed network. Second, while the current model integrates mRNA, miRNA, and DNA methylation data, it omits other valuable omics layers such as proteomics, metabolomics, and copy number variation, which could provide complementary biological insights. Third, we used an early integration strategy, where multi-omics data (mRNA, miRNA, and methylation) are concatenated into a single feature vector prior to graph modeling. While this approach simplifies representation learning, it may obscure modality-specific characteristics and interactions.

For future work, we plan to extend the framework by incorporating a broader range of multi-omics data and utilizing dynamic and context-specific interaction networks to enhance model performance and biological interpretability. In addition, implementing attention mechanisms to weigh the contributions of different omics features may further improve predictive accuracy. Also, we will explore late integration strategies, such as using modality-specific encoders followed by attention-based or gating fusion mechanisms, which may better capture complementary signals across omics layers and improve both performance and interpretability.

## 5. Conclusion

This study introduces MOGKAN, a novel deep learning framework for accurate and interpretable cancer classification using multi-omics data. The approach integrates a three-step data preprocessing pipeline, combining DESeq2, LIMMA, and LASSO regression to preserve key biological signals while reducing dimensionality. By fusing DNA methylation, miRNA, and mRNA data with Protein-Protein Interaction (PPI) networks, MOGKAN achieves a classification accuracy of 96.28% across 31 cancer types. Through the application of the Kolmogorov–Arnold theorem, the framework extracts hierarchical features that enhance both predictive performance and biological interpretability. Key biomarkers identified by MOGKAN including MCL1, GALNT6, and ITGB3 were validated through GO and KEGG pathway analyses, confirming their involvement in critical processes like PI3K-AKT signaling, lipid metabolism, and immune evasion. These findings demonstrate the capability of the proposed framework to uncover fundamental molecular drivers of cancer and support its potential for clinical application in personalized cancer therapy.

### Acknowledgments

We appreciate the assistance by Srikar Chittemsetty with the implementation of codes for experimental results.

## References

[1] Narrandes, S. & Xu, W. Gene expression detection assay for cancer clinical use. *J. Cancer* **9**, 2249 (2018).

[2] Singh, K. P., Miaskowski, C., Dhruva, A. A., Flowers, E. & Kober, K. M. Mechanisms and measurement of changes in gene expression. *Biol. Res. Nurs*. **20**, 369–382 (2018).

[3] Li, M., Sun, Q. & Wang, X. Transcriptional landscape of human cancers. *Oncotarget* **8**, 34534 (2017).

[4] Heo, Y. J., Hwa, C., Lee, G. H., Park, J. M. & An, J. Y. Integrative multi-omics approaches in cancer research: from biological networks to clinical subtypes. *Mol. Cells* **44**, 433–443 (2021).

[5] Menyhárt, O. & Győrffy, B. Multi-omics approaches in cancer research with applications in tumor subtyping, prognosis, and diagnosis. *Comput. Struct. Biotechnol. J*. **19**, 949–960 (2021).

[6] Geissler, K. et al. The role of aberrant DNA methylation in cancer initiation and clinical impacts. *Ther. Adv. Med. Oncol.* **16** (2024).

[7] Ankasha, S. J., Shafiee, M. N., Wahab, N. A., Ali, R. A. & Mokhtar, N. M. Post-transcriptional regulation of microRNAs in cancer: From prediction to validation. *Oncol. Rev.* **12** (2018).

[8] Chai, H. et al. Integrating multi-omics data through deep learning for accurate cancer prognosis prediction. *Comput. Biol. Med*. **134**, 104481 (2021).

[9] Bersanelli, M. et al. Methods for the integration of multi-omics data: mathematical aspects. *BMC Bioinformatics* **17**, 167 (2016).

[10] Ballard, J. L. et al. Deep learning-based approaches for multi-omics data integration and analysis. *BioData Min.* **17**, 38 (2024).

[11] Way, G. P. & Greene, C. S. Extracting a biologically relevant latent space from cancer transcriptomes with variational autoencoders. *Pac. Symp. Biocomput*. **2018**, 80–91 (2018).

[12] Kipf, T. N. & Welling, M. Semi-supervised classification with graph convolutional networks. Preprint at https://arxiv.org/abs/1609.02907 (2016).

[13] Hamilton, W. L. *Graph representation learning*. (Morgan & Claypool Publishers, 2020).

[14] Velickovic, P. et al. Graph attention networks. *Stat* **1050**, 10-48550 (2018).

[15] Xu, K., Hu, W., Leskovec, J. & Jegelka, S. How powerful are graph neural networks? Preprint at https://arxiv.org/abs/1810.00826 (2019).

[16] Kiamari, M., Kiamari, M. & Krishnamachari, B. GKAN: Graph Kolmogorov-Arnold Networks. Preprint at https://arxiv.org/abs/2406.06470 (2024).

[17] Zhang, F. & Zhang, X. GraphKAN: Enhancing Feature Extraction with Graph Kolmogorov Arnold Networks. Preprint at https://arxiv.org/abs/2406.13597 (2024).

[18] Ritchie, M. E. et al. LIMMA: Linear Models for Microarray and RNA-Seq Data Analysis, version 3.58.1. https://bioconductor.org/packages/limma (2024).

[19] Love, M. I., Huber, W. & Anders, S. DESeq2: Differential expression analysis based on the Negative Binomial distribution, version 1.42.0. https://bioconductor.org/packages/DESeq2 (2024).

[20] Zhang, F. & Zhang, X. GraphKAN: Enhancing Feature Extraction with Graph Kolmogorov Arnold Networks. Preprint at https://arxiv.org/abs/2406.13597 (2024).

[21] Kiamari, M., Kiamari, M. & Krishnamachari, B. GKAN: Graph Kolmogorov-Arnold Networks. Preprint at https://arxiv.org/abs/2406.06470 (2024).

[22] Bresson, R. et al. KAGNNS: Kolmogorov-Arnold Networks Meet Graph Learning. Preprint at https://arxiv.org/abs/2406.18380 (2024).

[23] Carlo, G. D., Mastropietro, A. & Anagnostopoulos, A. Kolmogorov-Arnold Graph Neural Networks. Preprint at https://arxiv.org/abs/2406.18354 (2024).

[24] Ahmed, T. & Sifat, M. H. GraphKAN: Graph Kolmogorov Arnold Network for Small Molecule-Protein Interaction Predictions. In ICML'24 *Workshop ML for Life and Material Science: From Theory to Industry Applications* (2024).

[25] Li, R., Li, M., Liu, W. & Chen, H. GNN-SKAN: Harnessing the Power of SwallowKAN to Advance Molecular Representation Learning with GNNs. Preprint at https://arxiv.org/abs/2408.01018 (2024).

[26] Yuan, L., Sun, S., Jiang, Y., Zhang, Q., Ye, L., Zheng, C. H., & Huang, D. S. scRGCL: a cell type annotation method for single-cell RNA-seq data using residual graph convolutional neural network with contrastive learning. *Briefings in Bioinformatics*, 26(1), bbae662 (2025).

[27] Yuan, L., Zhao, L., Jiang, Y., Shen, Z., Zhang, Q., Zhang, M., ... & Huang, D. S. scMGATGRN: a multiview graph attention network–based method for inferring gene regulatory networks from single-cell transcriptomic data. *Briefings in bioinformatics*, *25*(6), bbae526 (2024).

[28] Yuan, L., Xu, Z., Meng, B., & Ye, L. scAMZI: attention-based deep autoencoder with zero-inflated layer for clustering scRNA-seq data. *BMC genomics*, *26*(1), 350 (2025).

[29] Yuan, L., Zhao, L., Lai, J., Jiang, Y., Zhang, Q., Shen, Z., ... & Huang, D. S. iCRBP-LKHA: large convolutional kernel and hybrid channel-spatial attention for identifying circRNA-RBP interaction sites. *PLOS Computational Biology*, *20*(8), e1012399 (2024).

[30] Weinstein, J. N. et al. The cancer genome atlas pan-cancer analysis project. *Nat. Genet.* **45**, 1113–1120 (2013).

[31] Robinson, M. D., McCarthy, D. J. & Smyth, G. K. edgeR: a Bioconductor package for differential expression analysis of digital gene expression data. *Bioinformatics* **26**, 139–140 (2010).

[32] Ritchie, M. E. et al. limma powers differential expression analyses for RNA-sequencing and microarray studies. *Nucleic Acids Res*. **43**, e47 (2015).

[33] Tibshirani, R. Regression shrinkage and selection via the LASSO. *J. R. Stat. Soc. Series B Stat. Methodol*. **58**, 267–288 (1996).

[34] Rapaport, F. et al. Comprehensive evaluation of differential gene expression analysis methods for RNA-seq data. *Genome Biol*. **14**, 1–13 (2013).
[35] Chen, J. et al. An epigenome-wide analysis of socioeconomic position and tumor DNA methylation in breast cancer patients. *Clin. Epigenetics* **15**, 68 (2023).
[36] Pidsley, R. et al. A data-driven approach to preprocessing Illumina 450K methylation array data. *BMC Genomics* **14**, 1–10 (2013).
[37] Picard, M., Scott-Boyer, M. P., Bodein, A., Périn, O., & Droit, A. Integration strategies of multi-omics data for machine learning analysis. *Computational and Structural Biotechnology Journal*, *19*, 3735-3746 (2021).
[38] Z. Liu, Y. Wang, S. Vaidya, F. Ruehle, J. Halverson, M. Soljačić, T. Y. Hou and M. Tegmark, "Kan: Kolmogorov-arnold networks," arXiv preprint arXiv:2404.19756, 2024.
[39] Jensen, L. J. et al. STRING 8—a global view on proteins and their functional interactions in 630 organisms. *Nucleic Acids Res*. **37**, D412–D416 (2009).
[40] Szklarczyk, D. et al. The STRING database in 2021: customizable protein–protein networks, and functional characterization of user-uploaded gene/measurement sets. *Nucleic Acids Res*. **49**, D605–D612 (2021).
[41] Franceschini, A. et al. STRING v9.1: protein-protein interaction networks, with increased coverage and integration. *Nucleic Acids Res*. **41**, D808–D815 (2012).
[42] PyTorch Geometric. Graph Attention Convolution (GATConv) layer, version 2.4.0. https://pytorch-geometric.readthedocs.io/en/latest/generated/torch_geometric.nn.conv.GATConv.html (2024).
[43] Mostavi, M., Chiu, Y. C., Huang, Y. & Chen, Y. Convolutional neural network models for cancer type prediction based on gene expression. *BMC Med. Genomics* **13**, 1–13 (2020).
[44] Ramirez, R. et al. Classification of cancer types using graph convolutional neural networks. *Front. Phys.* **8**, 203 (2020).
[45] Kaczmarek, E. et al. Multi-omic graph transformers for cancer classification and interpretation. *Pac. Symp. Biocomput.* **2022**, 373–384 (2021).
[46] Lee, K. M. et al. MYC and MCL1 cooperatively promote chemotherapy-resistant breast cancer stem cells via regulation of mitochondrial oxidative phosphorylation. *Cell Metab.* **26**, 633–647 (2017).
[47] Sun, X. et al. GALNT6 promotes bladder cancer malignancy and immune escape by epithelial-mesenchymal transition and CD8+ T cells. *Cancer Cell Int.* **24**, 308 (2024).
[48] Chen, J. et al. CCL18 from tumor-associated macrophages promotes breast cancer metastasis via PITPNM3. *Cancer Cell* **19**, 541–555 (2011).
[49] Liu, F. & Wen, C. LINC01410 knockdown suppresses cervical cancer growth and invasion via targeting miR-2467-3p/VOPP1 axis*. Cancer Manag. Res*. 855–861 (2020).
[50] Chen, Y., Zhang, X., Li, J. & Zhou, M. Immune-related eight-lncRNA signature for improving prognosis prediction of lung adenocarcinoma. *J. Clin. Lab. Anal*. **35** (2021).
[51] Yan, Y., Liu, J., Xu, Z., Ye, M. & Li, J. lncRNA PCAT14 is a diagnostic marker for prostate cancer and is associated with immune cell infiltration. *Dis. Markers* **2021**, 9494619 (2021).
[52] Guo, W., *et al*. Single-exosome profiling identifies ITGB3+ and ITGAM+ exosome subpopulations as promising early diagnostic biomarkers and therapeutic targets for colorectal cancer. *Res.* **6**, 0041 (2023).
[53] Zhao, Q. *et al.* Comprehensive profiling of 1015 patients' exomes reveals genomic-clinical associations in colorectal cancer. *Nat. Commun.* **13**, 2342 (2022).
[54] Na, L. *et al.* MAML3 contributes to induction of malignant phenotype of gallbladder cancer through morphogenesis signalling under hypoxia. *Anticancer Res.* **43**, 2909–2922 (2023).
[55] Du, J., Wang, B., Li, M., Wang, C., Ma, T. & Shan, J. A novel intergenic gene between SLC8A1 and PKDCC-ALK fusion responds to ALK TKI WX-0593 in lung adenocarcinoma: a case report. *Front. Oncol.* **12**, 898954 (2022).
[56] Fruman, D. A., Chiu, H., Hopkins, B. D., Bagrodia, S., Cantley, L. C. & Abraham, R. T. The PI3K pathway in human disease. *Cell* **170**, 605–635 (2017).

[57] Broadfield, L. A., Pane, A. A., Talebi, A., Swinnen, J. V. & Fendt, S. M. Lipid metabolism in cancer: new perspectives and emerging mechanisms. *Dev. Cell* **56**, 1363–1393 (2021).

[58] Kanehisa, M., Furumichi, M., Sato, Y., Matsuura, Y. & Ishiguro-Watanabe, M. KEGG: biological systems database as a model of the real world. *Nucleic Acids Res*. 53, D672–D677 (2025).

[59] Kanehisa, M. Toward understanding the origin and evolution of cellular organisms. *Protein Sci*. 28, 1947–1951 (2019).

[60] Kanehisa, M. & Goto, S. KEGG: Kyoto Encyclopedia of Genes and Genomes. *Nucleic Acids Res*. 28, 27–30 (2000).

[61] Sies, H., Belousov, V. V., Chandel, N. S., Davies, M. J., Jones, D. P., Mann, G. E., Murphy, M. P., Yamamoto, M. & Winterbourn, C. Defining roles of specific reactive oxygen species (ROS) in cell biology and physiology. *Nat. Rev. Mol. Cell Biol.* **23**, 499–515 (2022).

[62] Brockhausen, I. Mucin-type O-glycans in human colon and breast cancer: glycodynamics and functions. *EMBO Rep.* **7**, 599–604 (2006).

[63] Hannun, Y. A. & Obeid, L. M. Sphingolipids and their metabolism in physiology and disease. *Nat. Rev. Mol. Cell Biol.* **19**, 175–191 (2018).

[64] Goffin, V., Binart, N., Touraine, P. & Kelly, P. A. Prolactin: the new biology of an old hormone. *Annu. Rev. Physiol.* **64**, 47–67 (2002).

[65] Gloerich, M. & Bos, J. L. Regulating Rap small G-proteins in time and space. *Trends Cell Biol.* **21**, 615–623 (2011).

[66] Hoxhaj, G. & Manning, B. D. The PI3K–AKT network at the interface of oncogenic signalling and cancer metabolism. *Nat. Rev. Cancer* **20**, 74–88 (2020).

[67] Spat, A. & Hunyady, L. Control of aldosterone secretion: a model for convergence in cellular signaling pathways. *Physiol. Rev.* **84**, 489–539 (2004).

[68] Gallagher, E. J. & LeRoith, D. Obesity and diabetes: the increased risk of cancer and cancer-related mortality. *Physiol. Rev.* **95**, 727–748 (2015).

[69] Yuan, L., Sun, S., Zhang, Q., Li, H. T., Shen, Z., Hu, C., ... & Huang, D. S. Identification of ferroptosis-related lncRNAs for predicting prognosis and immunotherapy response in non-small cell lung cancer. *Future Generation Computer Systems*, *159*, 204-220 (2024).

[70] Bao, W., Yang, B., & Chen, B. TAPE_selection: Organelle Proteins Classification with TAPE Feature Selection. *IEEE Transactions on Computational Biology and Bioinformatics* (2025).

[71] Chen, B., Li, N., & Bao, W. CLPr_in_ML: cleft lip and palate reconstructed features with machine learning. *Current Bioinformatics*, *20*(2), 179-193 (2025).